\pdfoutput=1

\documentclass[11pt]{article}

\usepackage{acl}

\usepackage{times}
\usepackage{latexsym}
\usepackage{amsmath}
\usepackage{booktabs}
\usepackage{graphicx}
\usepackage{makecell}

\usepackage{amsmath,amsfonts,bm}

\def\eqref#1{equation~\ref{#1}}

\def\1{\bm{1}}

\def\rvh{{\mathbf{h}}}

\DeclareMathAlphabet{\mathsfit}{\encodingdefault}{\sfdefault}{m}{sl}
\SetMathAlphabet{\mathsfit}{bold}{\encodingdefault}{\sfdefault}{bx}{n}

\def\gA{{\mathcal{A}}}

\def\gC{{\mathcal{C}}}

\def\gL{{\mathcal{L}}}

\def\gP{{\mathcal{P}}}

\def\gT{{\mathcal{T}}}

\def\gV{{\mathcal{V}}}

\newcommand{\eg}{\textit{e.g.}}

\usepackage[T1]{fontenc}

\usepackage[utf8]{inputenc}

\usepackage{microtype}

\title{Grounding Language Model with Chunking-Free In-Context Retrieval}

\author{Hongjin Qian$^1$, Zheng Liu$^2$$^*$, Kelong Mao$^1$, Yujia Zhou$^1$, Zhicheng Dou$^1$\thanks{$^*$Corresponding author.} \\
        $^1$ Gaoling School of Artificial Intelligence, Renmin University of China \\ 
        $^2$ Beijing Academy of Artificial Intelligence \\
        \texttt{\{ian,dou\}@ruc.edu.cn} \quad
        \texttt{zhengliu1026@gmail.com} \\
}

\begin{document}
\maketitle

\begin{abstract}

This paper presents a novel Chunking-Free In-Context (CFIC) retrieval approach, specifically tailored for Retrieval-Augmented Generation (RAG) systems. Traditional RAG systems often struggle with grounding responses using precise evidence text due to the challenges of processing lengthy documents and filtering out irrelevant content. Commonly employed solutions, such as document chunking and adapting language models to handle longer contexts, have their limitations. These methods either disrupt the semantic coherence of the text or fail to effectively address the issues of noise and inaccuracy in evidence retrieval.

CFIC addresses these challenges by circumventing the conventional chunking process. It utilizes the encoded hidden states of documents for in-context retrieval, employing auto-aggressive decoding to accurately identify the specific evidence text required for user queries, eliminating the need for chunking. CFIC is further enhanced by incorporating two decoding strategies, namely Constrained Sentence Prefix Decoding and Skip Decoding. These strategies not only improve the efficiency of the retrieval process but also ensure that the fidelity of the generated grounding text evidence is maintained.
Our evaluations of CFIC on a range of open QA datasets demonstrate its superiority in retrieving relevant and accurate evidence, offering a significant improvement over traditional methods. By doing away with the need for document chunking, CFIC presents a more streamlined, effective, and efficient retrieval solution, making it a valuable advancement in the field of RAG systems.                 

\end{abstract}

\section{Introduction}
\label{sec:intro}
Recently, retrieval-augmented generation (RAG) has marked a significant advancement in the field of natural language processing (NLP). This technique has demonstrated remarkable effectiveness in reducing hallucination in text generation~\cite{ji2023survey}, particularly in knowledge-intensive tasks like open-domain question answering~\cite{wang-etal-2019-multi,lewis2020retrieval,shuster-etal-2021-retrieval-augmentation,komeili-etal-2022-internet}.  An RAG system typically consists of two components: the retriever and the generator. Given an input query, the retriever first identifies relevant evidence text, upon which the generator then generates the answer.

Ideally, the generator's output should be grounded by precise evidence text obtained by the retriever. However, this poses a challenge for most retrieval systems, as they often retrieve lengthy documents such as web pages. However, in practice, we only need specific grounding text from these documents to help answer user queries. Using lengthy documents directly in the RAG system presents two difficulties. First, generation models may struggle to handle the extensive length of these documents. Second, irrelevant or distracting content within the documents can lead the model astray from the main query, resulting in inaccurate response generation \citep{shi2023large}.
            
\begin{figure}
    \centering
    \includegraphics[width=\linewidth]{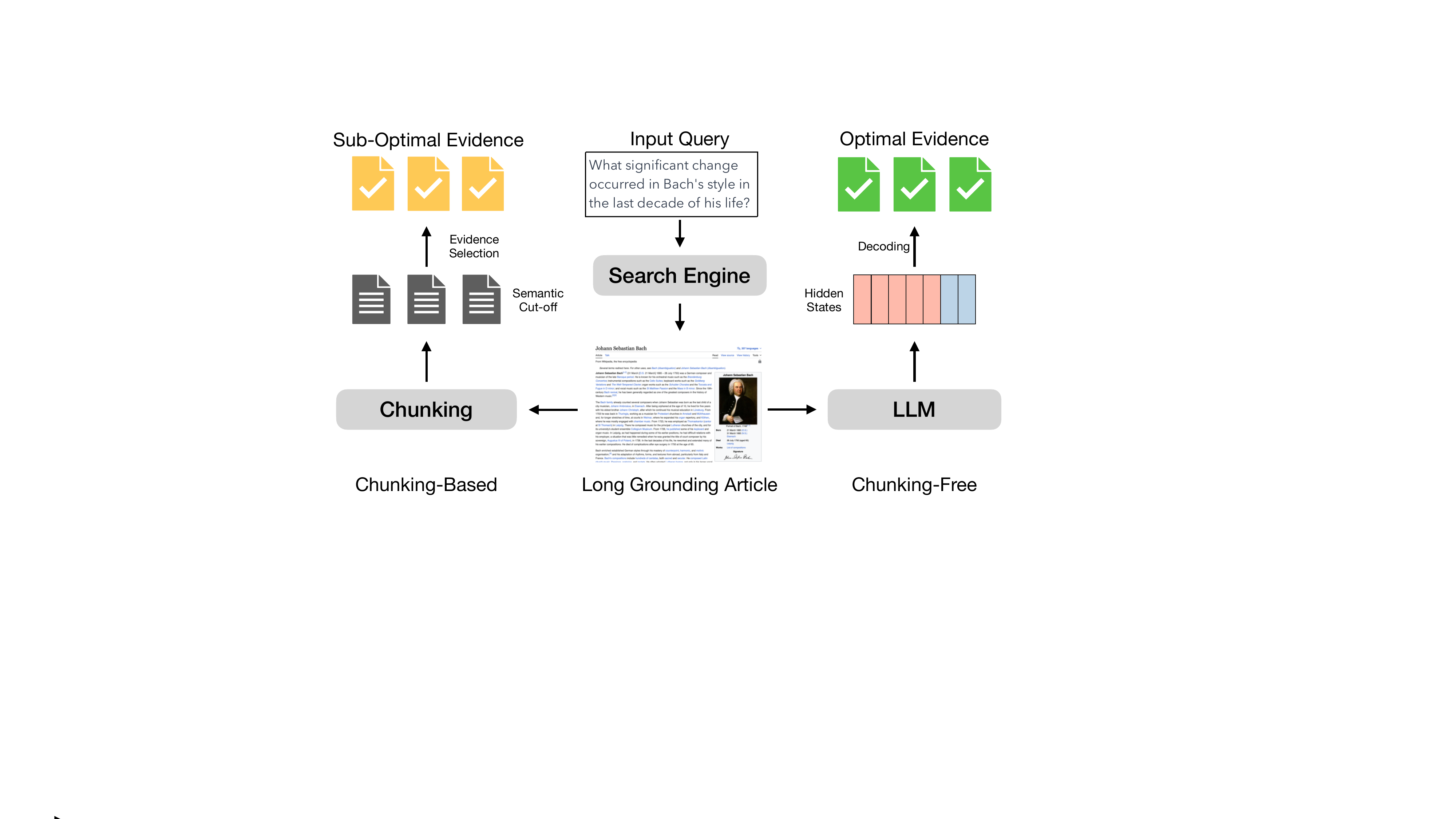}
    \caption{Comparison of Chunking-Based and Chunking-Free Methods. The left panel illustrates the chunking-based method, involving chunking a lengthy document into smaller passages followed by refinement through passage ranking. The right panel depicts the chunking-free method proposed in this paper, where grounding text is directly decoded by LLMs without the need for document chunking.}
    \label{fig:idea}
    \vspace{-20pt}
\end{figure}

To address this issue, common approaches involve chunking documents into smaller passages and employing strategies like reranking for relevance~\citep{nogueira2020passage,mao-etal-2021-reader,gao2024retrievalaugmented}, or selecting passages based on other measurements~\cite{asai-etal-2022-evidentiality, jiang2023active,qian2023optimizing}. However, the chunking process is often sub-optimal, as determining the granularity of the passage chunking is challenging. Improper chunking can disrupt the semantics and result in incomplete and incoherent retrieved information~\cite{dong2023survey}.
Another method involves adapting large language models~(LLMs) to process longer contexts by training them on long contexts or implementing a sliding context window~\cite{pcw,longctx,longlora}. While these methods enable LLMs to handle longer texts, they do not fully address the issue of noise in the lengthy documents and cannot output the grounding text for the generated response~\cite{kaddour2023challenges}. 

In this paper, we propose a Chunking-Free In-Context~(CFIC) retrieval approach aimed at helping the RAG system mitigate information bias introduced by document chunking and irrelevant noisy text. Specifically, given an input query and a long grounding document, instead of refining the long documents with a chunking-based method, we leverage the document's encoded hidden states to perform Chunking-free In-Context Retrieval. It circumvents the traditional chunking process, allowing the retrieval system to auto-aggressively decode and pinpoint the precise evidence text to ground the response generation to a query. Figure \ref{fig:idea} shows the comparison between the chunking-based method and the chunking-free method for grounding text retrieval. The chunking-free method demonstrates a superior ability to identify optimal evidence text, as it considers the entire document for a comprehensive perspective.  

Concretely, CFIC involves encoding a document into transformer hidden states. When a user query is input, CFIC continues to encode the query alongside task instructions following the hidden states, subsequently generating grounding text. In practice, we can cache the documents' hidden states to further reduce computation\footnote{In a single-sided transformer model, the forward side is auto-regressive; once an output token's hidden state is computed, it remains unchanged for subsequent forward steps, allowing us to use these encoded states as a cache.}. Given the expectation for CFIC to process lengthy documents, it becomes imperative to adapt CFIC for handling long contexts. Considering the trade-off between efficiency and effectiveness, in this paper, we adapt CFIC to accommodate a 32k context, utilizing LLAMA2-7B-chat as the foundational model. To achieve this, we construct a dataset containing long document, user query and precise text evidence to training the foundation model via Supervised Fine-Tuning~(SFT).

Despite its promise, CFIC encounters two major challenges: (1) \textbf{Efficiency issue}: the auto-aggressive generation process involves executing attention interactions for generating each new token, a procedure that becomes particularly time-consuming with longer contexts due to the management of exponentially larger attention matrices. This process requires substantial computational resources.~\cite{kaplan2020scaling}, and (2) \textbf{Faithfulness issue}: it is challenging to ensure the generation model's output remains faithful to the original input context, given its open-ended decision boundary~\cite{li2022faithfulness}. To address these, we propose two decoding strategies that accelerate inference and ensure that generated text evidence originates from the corpus. These include: (1) utilizing sentence prefixes as decoding candidates to shift the model's decision boundary from open-ended to document-dependent generation and (2) upon locating the appropriate sentence prefix, bypassing the decoding of intermediate tokens and directly selecting sentence ends with the highest likelihood of the $[\text{eos}]$ token, thereby terminating the generation. Furthermore, to retrieve multiple text spans as evidence, we sample several sentence prefixes with the best likelihood as candidates and rank them by sequence likelihood. 
By this means, CFIC not only enhances the relevance and accuracy of retrieved evidence text but also preserves the semantic integrity of the information, effectively addressing major drawbacks of current retrieval systems. 

We tested CFIF on the LongBench tasks~\cite{bai2023longbench} including: (1) single-document question answering with datasets like NarrativeQA, Qasper, MulitfieldQA, and (2) multi-document QA with datasets like Musqus and HotpotQA. The experiment results verify the effectiveness of our method. In summary, our contributions are as follows: (1) we propose a chunking-free in-context retrieval method dedicated to the RAG system, aiding in locating precise text evidence to answer user queries;  (2) we propose the CFIC model of which the ability to find text evidence from long context is enhanced via Supervised Fine-Tuning with self-constructed dataset; (3) we design two decoding strategies that significantly improve the efficiency and accuracy of the CFIC's decoding process.











\section{Related Work}
The RAG framework, initially introduced in the works of \citet{lewis2020retrieval}, aimed to enhance language models' capacity for generating knowledge-based responses\cite{FiD, chung2022scaling, AutoGPT}. Subsequent research primarily focus on refining the RAG's two core components. On the retrieval front, significant strides have been made towards more efficient and precise retrieval methods~\cite{kNN-LM,ease,li2022large,sugre,guo22webformer}. For example,  the arise of Dense Passage Retrieval significantly surpasses traditional sparse dense~\cite{karpukhin-etal-2020-dense}. Parallel efforts on the generation side have concentrated on fine-tuning generative models to better harmonize with retrieved information, a notable example being the work of~\citet{izacard2021leveraging} in optimizing external knowledge utilization~\cite{izacard2021distilling, chung2022scaling, kamalloo2023evaluating}.

Nevertheless, RAG encounters specific challenges, especially in managing lengthy and complex retrieved documents. Researchers, including \citet{mao-etal-2021-reader}, have developed chunking and reranking techniques to enhance passage relevance. Furthermore, \citet{guu2020realm} introduced methods for jointly learning retriever and generator models, thereby improving the coherence and relevance of outputs. Addressing the issue of lengthy contexts in RAG has involved either refining contexts~\cite{li2022large, jiang2023active} or adapting generation models to handle extended contexts~\cite{longctx, pcw,longlora}.

Recent advancements in RAG predominantly incorporate large-scale language models (LLMs), such as GPT-3 and GPT-4, to augment language processing capabilities~\cite{gpt-3,gpt-4,llama,gemini}. The integration of LLMs has paved the way for more contextually rich and nuanced generation, especially in aligning generated responses with human preferences~\cite{ICRALM, WebGLM}. In RAG systems employing LLMs, the accuracy of retrieved textual evidence is crucial for reducing hallucinations and incorporating external knowledge~\cite{raghallucination,hallucination,llmlie,multitaskhallu,mao2023large}. However, the challenge of processing long and noisy contexts persists~\cite{liu2023lost, li2022large, xu2023retrieval,qian2023webbrain}.
This paper introduces a chunking-free in-context retrieval approach that leverages transformer hidden states to generate grounding text evidence, treating evidence retrieval as a generative process. This method represents a more streamlined and efficient retrieval solution for RAG systems, marking a significant advancement over previous retrieval methodologies.

\section{Method}
\begin{figure*}[t]
    \centering
    \includegraphics[width=\linewidth]{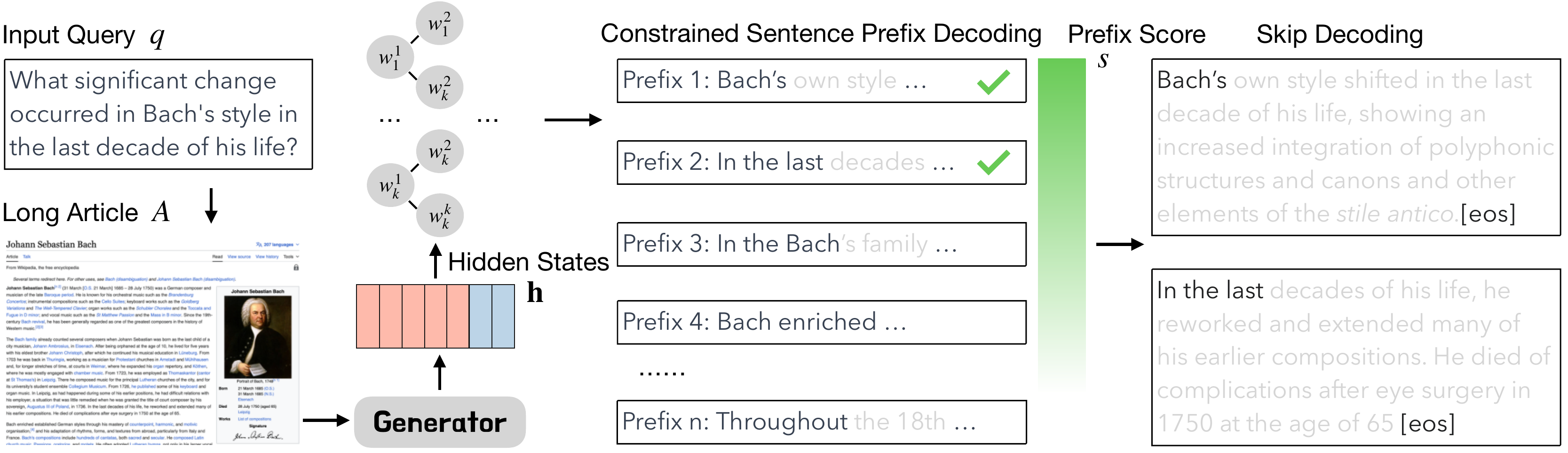}
    \caption{Overview of the proposed method: CFIC. The middle part shows the Constrained Sentence Prefix Decoding strategy which ensures the generated text prefixes originate from the input article. The right part shows the Skip Decoding strategy which bypasses decoding the intermediate tokens while terminating generation at the position with the best likelihood of [eos] token. \textcolor{gray}{Gray} tokens in the figure are bypassed during generation.}
    \label{fig:model}
    \vspace{-10pt}
\end{figure*}
\subsection{Preliminary}

In a RAG system, the system takes a user query $q$ as input, retrieves text evidence $\gA$ from a text corpus $\gC$ using a retriever $\theta(\cdot)$ as external knowledge, and utilizes a generation model $\phi(\cdot)$ to produce the final response $\gT$. This pipeline can be formalized as:
\begin{align}
\label{eq:def}
\gA = \theta(q,\gC), \quad
\gT = \phi(q,\gA).
\end{align}
The retriever $\theta(\cdot)$ can be either a standalone retriever (\eg, DPR~\cite{karpukhin-etal-2020-dense}) or a commercial search engine (\eg, Google), and the generation model $\phi(\cdot)$ is usually a trained LM. Based on Eq. (\ref{eq:def}), the quality of the generated text $\gT$ is bounded by the accuracy of the evidence $\gA$, emphasizing the importance of accurately finding the accurate text evidence.

In practice, most RAG systems' retrievers cannot accurately find exact text evidences, but only retrieve lengthy documents~(\eg, web pages or pre-indexed articles) that contain the evidences. As mentioned in Section~\ref{sec:intro}, such lengthy documents might bias the generated content.  
Thus, given the retrieved evidence $\gA$, we usually select a few useful text spans, called supporting text evidence $\gP=\{p_1,\cdots, p_k \} \in \gA$, to support the answer generation for the input query $q$ in a RAG system.

We define the process of finding supporting passages as a mapping function $f(\cdot)$:
\begin{align}
\gP=\{p_1,\cdots, p_k \} = f(\gA).
\label{eq:map}
\end{align}
The mapping function $f(\cdot)$ can take various forms, such as chunking the text evidence $\gA$ and prioritizing relevant chunks through re-ranking. In this paper, we define the mapping function $f(\cdot)$ as a generation process in which we directly generate the supporting text evidence $\gP$ conditioned on the transformer hidden-states $\rvh=\texttt{Trans}(\gA)$ of the lengthy document:
\begin{align}
    \gP=f(\gA) \sim \texttt{Generator}(\gP|\rvh, q).
    \label{eq:gen}
\end{align}

Compared to regular auto-regressive decoding, the above process is characterized by the fact that the generation target $\gP$ contains text sourced from $\gA$. This means that once we determine the decoding prefix, we can skip the intermediate tokens and directly find the terminating position by computing the probability of inserting $[\text{eos}]$ token. This greatly improves inference efficiency while ensuring that the generated text accurately represents the source text. Additionally, a single supporting passage may not always be sufficient for question answering. Therefore, we can obtain multiple sentence prefixes as top-$k$ candidates using sampling decoding. In this paper, our proposed model CFIC applies these ideas to generate the top-$k$ supporting text evidence $\gP$, which are further discussed in the following sections.

\subsection{The Proposed Model: CFIC}
Figure \ref{fig:model} presents an overview of our proposed model, CFIC. The process begins with CFIC receiving a user query. It then retrieves a long article as grounding evidence through a search engine (e.g., Google). Subsequently, CFIC combines the long document and the query into an input prompt, following the format outlined in Table \ref{tab:template}. This input prompt is encoded into hidden states. Based on these hidden states, CFIC first identifies the top-$k$ sentence prefix candidates using the Constrained Sentence Prefix Decoding strategy. This strategy ranks the sentence prefixes considering the generation score~(accumulated token log probabilities normalized by token length) of each sentence prefix. CFIC then skips the decoding of intermediate tokens and terminates the generation process by locating the [eos] token position with the highest likelihood (Skip Decoding). Consequently, we obtain $k$ grounding evidence texts that can aid in supporting downstream tasks.
It is important to note that this paper primarily focuses on pinpointing precise grounding text evidence within the long document, rather than on the retrieval of the long document. Therefore, we assess our CFIC and all baseline models using the LongBech benchmark, which provides pre-prepared long documents. In the subsequent sections, we will introduce the two proposed decoding strategies and then discuss the training and inference processes of CFIC.

\paragraph{Constrained Sentence Prefix Decoding}
Normally, the generation process of an auto-aggressive decoding model is as:
\begin{align}
    w_n \sim \prod_{n=1}^{|w|}p(w_n\in \gV|w_{<n},\rvh),
    \label{eq:agg}
\end{align}
where $\rvh$ represents the hidden states of previous tokens. The current token, denoted by $w_n$, is selected from the entire vocabulary $\gV$ of the generation model. In the case of CFIC, the generation target $\gP$ consists of text spans that originate directly from the source context. Consequently, it is possible to define a more constrained generation space to ensure the faithfulness of the text produced. Specifically, we suggest employing the prefix of each sentence within the source context as generation constraints. This approach guarantees that the text generated by CFIC can be traced back to the input context.
Thus, Eq.~(\ref{eq:agg}) can be modified as:
\begin{align}
    w_n \sim \prod_{n=1}^{|w|}p(w_n\in \bar{\gV}|w_{<n},\rvh),
    \label{eq:constrained}
\end{align}
where $\bar{\gV}$ denotes a token set contains each sentence's prefix. 

The sentence prefix serves as an position identifier to facilitate the identification of the starting point of a supporting passage within the source context. To select the top-$k$ candidate passages, it is essential to differentiate $k$ distinct sentence prefixes. This is achieved through the constrained top-$k$ sampling decoding, a process that entails selecting the next token $w_n$ from the top-$k$ most likely tokens $\bar{\gV}_k \in \bar{\gV}$ based on the token's probability, $p(w_n|w_{<n})$. The sampling process terminate once the generated sentence prefixes are capable of uniquely identifying positions in the source context. The number of decoding steps required until termination is denoted by $\beta$, resulting in up to $k^\beta$ prefix candidates after $\beta$ steps. We denote the generated sentence prefix by $b$. Subsequently, these prefix candidates are ranked according to the prefix sequence score $s$, which calculates the normalized accumulated log probability of tokens as follows:
\begin{equation}
    s = \frac{1}{|w|}\sum_{n=1}^{|w|}\log p(w_n|w_{<n}).
\end{equation}
Finally, the $k$ sentence prefixes with the highest scores are selected. 

Referring to Figure~\ref{fig:model} for illustration, the decoding process initiates by sampling $k$ tokens, such as [\textit{Bach, In, ..., Throughout}], to represent the first set of candidate tokens. Given that multiple sentences in the long article begin with the tokens [\textit{Bach, In}], the decoding of subsequent tokens is necessary. For sentences that start with "\textit{Bach}", the decoding terminates at step $\beta=2$. And for sentences beginning with "\textit{In}", the decoding ends at step $\beta=3$. Following this, we retain $k=2$ sentence prefixes to identify the supporting passages. 

\begin{table*}[t]
\small
    \centering
    \begin{tabular}{cccccccc}
    \toprule
         Dataset &  SFT  & NarrativeQA & Qasper & MultiFieldQA & HotpotQA & MuSiQue\\
    \midrule
         Num of Samples & 25,652  & 200 & 200 & 150 & 200 & 200\\
         Ave. Length &  12,248  & 18,409 & 3,619 & 4,559 & 9,151 & 11,214\\

    \bottomrule
    \end{tabular}
    \caption{Statistical information of the datasets utilized in this paper, where the average length indicates the word count, typically smaller than the BPE-tokenized token length.}
    \label{tab:data}
    \vspace{-10pt}
\end{table*}

\paragraph{Skip Decoding}

Similarly, since the generation target originates exactly from the source text, once the generation prefix is determined, we can use the generated prefix as a position identifier to locate the original text in the source text. Subsequently, we can bypass decoding the intermediate tokens and directly compute the token probability $p(\text{[eos]})$ for the $[\text{eos}]$ token after each sentence following the generated prefix. We select the position with the highest probability as the termination point. In practice, we calculate $p_{[\text{eos}]}$ after each sentence within a predefined token distance $d$. Formally, given a generated prefix $b$, we determine the termination position as follows: 
\begin{align}
w^*_{[\text{eos}]} = \arg\max_{l \in \gL} p_{[\text{eos}]}(b \oplus l), \quad
|l| \leq d,
\label{eq:skip}
\end{align}
where $l$ represents the token sequence following the prefix $b$ with a maximum length of $d$. 

\begin{table}
\centering

\small

\begin{tabular}{l}\\\toprule  
\makecell[l]{\textit{Below is an article, read the article }\\ \textit{and answer my question after the article.}}\\  
\makecell[l]{Now the article begins: \\ 
\texttt{\{Article\}} \\
Now the article ends.
} \\
\makecell[l]{
\textit{Select several sentences from the article }\\
\textit{to answer my question.}}\\
\makecell[l]{Question: \texttt{\{Question\}}} \\
\bottomrule
\end{tabular}
\caption{Prompt template in training and evaluation.\label{tab:template}}
\vspace{-10pt}
\end{table} 
\paragraph{Training and Inference}

As previously discussed, we define the task of identifying supporting passages from a long source text for grounding downstream tasks as evidence generation. To this end, it is crucial to enhance the generation model with the capability to pinpoint precise textual evidence within extensive texts. In this study, CFIC achieves this through Supervised Fine-Tuning (SFT). We employ a prompt, formed using the pair $(q,\gA)$ as outlined in Table \ref{tab:template}, as the input, and use the text evidence $\gP$ as the target for generation. The model is trained using the negative log-likelihood (NLL) loss function:
\begin{align}
    \mathcal{L}(q, \gA,\gP^*) = -\sum_{n=1}^{|\gP^*|}\log p(\gP^*_n|\gP^*_{<n},q,\gA).
\end{align}
The training dataset is introduced in Section \ref{sec:data}.

During the inference stage, given the input $(q,\gA)$, we apply Constrained Sentence Prefix Decoding and Skip Decoding strategies to extract $k$ supporting passages. Should these passages exhibit overlapping sections, we amalgamate such intersecting passages into a single cohesive passage. Subsequently, these collated supporting passages are utilized to ground downstream tasks.

\section{Experiment}

\subsection{Datasets and Evaluation Metric}
\label{sec:data}
As mentioned above, we train the CFIC model using data that contains $(q,\gA,\gP)$ triplets via SFT. Most current datasets cannot provide such data format. Thus, we use self-constructed SFT data to train the CFIC model, and evaluate all baselines on the LongBench benchmark~\cite{bai2023longbench}. Specifically, to construct the SFT training data, we first collect a corpus of lengthy articles, including Wikipedia articles, novels, and news articles. Subsequently, we randomly select text spans from these articles and ask ChatGPT to generate a query that can be answered by each text span. As for evaluation, we choose five datasets from LongBench including NarrativeQA~\cite{kočiský2017narrativeqa}, Qasper~\cite{dasigi2021dataset}, MultiFieldQA~\cite{bai2023longbench}), HotpotQA~\cite{yang2018hotpotqa} and MuSiQue~\cite{trivedi2022musique}. Following the LongBench benchmark, we use F1-score as the evaluation metric. For further details of LongBench, please refer to \citet{bai2023longbench}. We show the statistical information of all datasets in Table \ref{tab:data}.

\subsection{Baseline Settings}
In this study, we focus on in-context retrieval within the Retrieval-Augmented Generation (RAG) system. As such, we employ stand-alone LLMs as generators. Specifically, we utilize Llama2-7B-chat-4k~\cite{touvron2023llama} and Vicuna-v1.5-7B-16k~\cite{zheng2023judging} as our generators. To assess our chunking-free approach against the traditional chunking-based methods, the baseline model settings are as follow:


\paragraph{Chunking-Base Method}
Chunking-based methods generally commence by segmenting a lengthy document into smaller passages using heuristic strategies, followed by reranking these passages with a ranking model. In our research, we investigate two prevalent chunking strategies: (1). Sliding Window Chunking (SW): This strategy involves dividing the document into sentences and then grouping these sentences into passages. Each passage is designed not to exceed a predefined maximum length of 256 words, with a stride of one sentence. (2). Paragraph-based Chunking (Para): Here, the document is split by paragraph markers (e.g., \textbackslash n). We employ ``bge-large-en-v1.5''~\cite{bge_embedding} and ``llm-embedder''~\cite{llm_embedder} as the ranking models. We utilize the SW and Para strategies to divide the document into passages, which are then reranked by the ranking models. The highest-ranking passages are chosen as the input context for the generators to support the QA tasks.

\paragraph{Chunking-Free Method}
For the chunking-free models, we present the outcomes using Vicuna-v1.5-7B-16k~\cite{zheng2023judging}, LongChat-7B-32k~\cite{longchat2023}, and LongAlpaca-7B-32k~\cite{longlora} as baseline models. These models refine lengthy documents into concise text evidence, which then serves as context for generator to support QA tasks. To ensure a fair comparison, all baseline models provide a comparable volume of textual evidence for downstream tasks, maintaining consistency in the number of passages or token length. We also explore the effectiveness of feeding full articles into generators. We introduce our inplementation details in Appendix~\ref{sec:imp}.

\subsection{Main Results}

\begin{table*}[t]
    \centering
    \small
    
    \begin{tabular}{lccccccccccccccccccc}
    \toprule
    & &   \multicolumn{5}{c}{\textbf{Llama2-7B-chat-4k}} & \multicolumn{5}{c}{\textbf{Vicuna-v1.5-7B-16k}} \\ 
        Model & chunk &  nar & qas  & mul  & hot & mus & nar & qas  & mul  & hot & mus \\

    \midrule
        BGE  & SW          & 13.9 & 22.0 & 34.0 & \underline{34.0} & 14.0 & 12.1 & 27.3 & 37.5 & \underline{33.6} & \underline{13.5}  \\
        BGE  & Para        & 12.1 & 21.7 & 31.4 &  31.2 & 12.3 & 10.2 & 23.2 & 34.7 &  31.7 & 12.5&    \\
        LLM-Embedder  & SW &  14.1 & \underline{23.2} & 34.3 &  33.8 & \underline{14.6} & 13.2 & \underline{27.4}& \underline{39.1} &  31.6 & 12.6  \\
        LLM-Embedder & Para &  13.2 & 21.7 & 34.1 & 32.9 & 12.6 & 12.3 & 25.1 & 36.3 &  31.1 & 12.1  \\
    \midrule
        Vicuna-7B & - & 13.7 & 19.0 & 23.3 &  22.0 & 9.7 &12.3 & 23.5 & 24.0 &  23.8 & 11.0 \\
        LongChat-7B & - & 12.2 & 19.7 &29.5 &  27.9 & 9.6 & 11.1 &21.9 &32.4 &30.2&9.7  \\
        LongAlpaca-7B  & - &  12.8 & 19.3 & 26.8  & 28.8 & 10.3 & 11.2 & 21.2 & 25.2 & 27.2 & 10.2 \\ 
        CFIC-7B(Ours)& - & \underline{18.3} & \textbf{27.7} & \textbf{41.2} & \textbf{34.0} & \textbf{14.7} & \underline{17.5} & \textbf{31.0} & \textbf{39.8}  & \textbf{33.8} & \textbf{16.2}  \\
        \midrule
        Full Article & - & \textbf{18.7} & 19.2 & \underline{36.8} &  32.8 & 9.4 &\textbf{19.4}  &26.1  & 38.5& 25.3 &  9.8\\
        
    \bottomrule
    
    \end{tabular}
    \caption{Main experiment results, which are the QA performance across various datasets, using different refined text evidence as context.  Following~\citet{bai2023longbench}, we use F1-score as the evaluation metric.
    The best results are in bold and the secondary results are marked with underline.}\label{tab:overall}
\end{table*}

Table \ref{tab:overall} shows the main experiment results which are the performance across different QA tasks using various refined text evidence as context. From the results we have the following findings:
\textbf{First}, CFIC significantly outperforms other LLMs in chunking-free in-context retrieval tasks as CFIC is specifically optimized to select precise text evidence crucial for grounding QA tasks. This underscores the necessity and effectiveness of supervised fine-tuning (SFT) in adpting the foundation model into the in-context retrieval task.
\textbf{Second}, Chunking-based methods serve as strong baselines due to their ability to extract passages directly from the source context, whereas LLMs lacking SFT tend to generate content that may not always align faithfully with the source material. CFIC, however, consistently surpass all chunking-based baselines across all datasets, indicating the potentiality of the chunking-free in-context retrieval paradigm.
\textbf{Last}, Compared to using the entire article as context, our CFIC model significantly improves the performance of QA tasks across most datasets, except for the NarrativeQA dataset. This improvement evidences the critical role of identifying and utilizing the right and precise context in optimizing QA task performance, demonstrating the CFIC model's efficiency in context filtering and utilization. As for the NarrativeQA dataset, we find that NarrativeQA's precise text evidence frequently appears at the start of lengthy articles, a location that LLMs tend to prioritize their attention~\cite{liu2023lost}. This might explain why CFIC does not perform as well on this dataset, given that its approach to identifying precise evidence could inadvertently introduce errors, thereby diminishing its accuracy. In practice, however, that precise text evidence can be located throughout the entire length of an article, not just at the beginning.

\subsection{Discussion}
\paragraph{Ablation Study}
To assess the effectiveness of the design of CFIC, we conduct an ablation study by removing key components of the model, including: 
(1). Removal of Sentence Prefix Decoding Strategy (\textit{w/o} prefix): we remove the constraint of limiting the decoding space to sentence prefixes. Instead, a beam search algorithm was employed to sample short sequences (each comprising 8 tokens) based on the input article. Subsequently, the top-$k$ short sequences were matched back to the input article to identify starting prefixes.
(2). Removal of Skip Decoding (\textit{w/o} skip): we dispensed with the practice of bypassing intermediate tokens following the sentence prefix decoding. The model continued to decode the remaining tokens up to a maximum length of 256 tokens.
(3). Removal of Both Decoding Strategies (\textit{w/o} both): the CFIC model was tasked to decode outputs using a greedy search algorithm, devoid of both the sentence prefix and skip decoding strategies.
(4). Absence of SFT (LongAlpaca-7B): LongAlpaca-7B is a context-extended version of LLAMA2-7B-chat. We utilized LongAlpaca-7B as the base model, representing the variant of CFIC without task-specific SFT.

The results of the ablation experiments are presented in Table \ref{tab:abl}. Our findings can be summarized as follows:
(1). The removal of any of the CFIC model components resulted in a notable degradation in performance, underscoring the collective contribution of these elements to the model's effectiveness.
(2). The most substantial decrease in performance was observed when SFT was omitted. This suggests that the vanilla LLM struggles to accurately locate precise grounding text from lengthy documents, despite its enhanced capability for processing extended contexts.
(3). Removing either the sentence prefix decoding or the skip decoding strategies led to an obvious reduction in performance. This finding verifies our hypothesis that these decoding strategies not only curtail decoding computational demands but also improve the fedelity of the generated grounding text.

\begin{table}[t]
    \centering
    \small
    
    \begin{tabular}{lccccc}
    \toprule
   &  \multicolumn{5}{c}{\textbf{Llama2-7B-chat-4k}}   \\
  Model  &  nar & qas  & mul  & hot & mus \\
    \midrule
        CFIC-7B & \underline{18.3} & \textbf{27.7} & \textbf{41.2} & \textbf{34.0} & \textbf{14.7}\\
        \quad \textit{w/o} prefix & 16.4 & 26.0 & \underline{39.3} & \underline{33.0} & \underline{12.5}\\
        \quad \textit{w/o} skip  & 15.8 & 27.0 & 37.6 & 30.1 & 11.6\\
        \quad \textit{w/o} both & 13.2 & 20.2 & 37.4 & 30.1 & 9.2\\
        LongAlpaca-7B &  12.8 & 19.3 & 26.8  & 28.8 & 10.3 \\

        \midrule
        Full Article  & \textbf{18.7} & 19.2 & 36.8 &  32.8 & 9.4 \\
        
    \bottomrule
    
    \end{tabular}
    \caption{Results of the ablation Study. }
    \label{tab:abl}
    \vspace{-10pt}
\end{table}
\begin{table*}[t]
\small
    \centering
    
    \begin{tabular}{p{.98\linewidth}}
    \toprule
     \textbf{Query}: What hedge fund's collapse in 1998 highlighted the need for regulation of derivatives? \\
     \textbf{Answer}: Long Term Capital Management (LTCM)  \\
\midrule
    \textbf{CFIC-7B}: \textcolor{teal}{In 1998, a trillion-dollar hedge fund called Long Term Capital Management (LTCM) was near collapse.} Using mathematical models to calculate debt risk, LTCM used derivatives to leverage \$5 billion into more than \$1 trillion.
    \textcolor{teal}{The derivative transactions were not regulated}, nor were investors able to evaluate LTCM's exposures. \\
\midrule
     \textbf{LongAlpaca-7B}:  The catastrophic financial events of recent months have proved them (Born and Sheila Bair) right.
      In 2010, a documentary film Inside Job further alleged that \textcolor{teal}{derivatives regulation was ineffective} from the Clinton administration on. \\
\midrule
     \textbf{GPT-3.5-Turbo}: The hedge fund whose collapse in 1998 highlighted the need for regulation of derivatives was \textcolor{teal}{Long Term Capital Management (LTCM)}.
 \\
\midrule
    \textbf{GPT-4}: \textcolor{teal}{In 1998, a trillion-dollar hedge fund called Long Term Capital Management (LTCM) was near collapse.} Using mathematical models to calculate debt risk, LTCM used derivatives to leverage \$5 billion into more than \$1 trillion. \\
    \bottomrule
    \end{tabular}
    \caption{Results of Case Study. The text colored with \textcolor{teal}{teal} refers to the grounding evidence for the user query.}
    \vspace{-10pt}
    \label{tab:case}
\end{table*}   
\paragraph{Choice of Decoding Length}
In our CFIC model, as defined in Eq.~(\ref{eq:skip}), the generation process is terminated upon locating the position of the [eos] token within a predetermined distance $d$. This distance is analogous to the maximum generation length typically set in standard text generation tasks, which governs the length of the decoded text. The selection of $d$ involves a careful balance: too small a value may lead to excessively brief output grounding text, offering scant information for substantiating downstream tasks. Conversely, a larger $d$ may result in longer output texts, potentially introducing additional textual noise and necessitating increased computational resources to process the extended sequences.

\begin{figure}
    \centering
    \includegraphics[width=\linewidth]{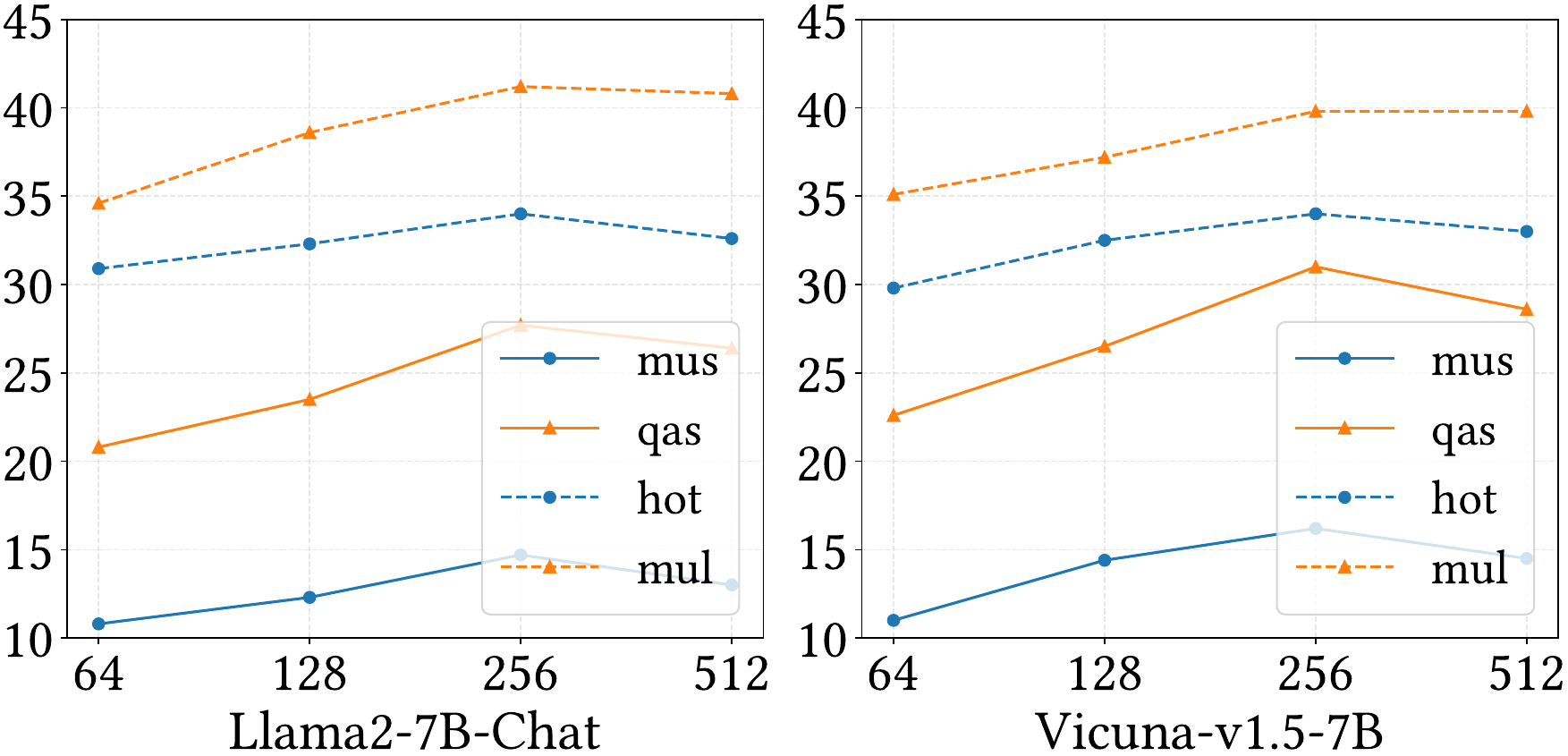}
    \caption{The choice of Maximum Decoding Length.}
    \label{fig:window}
    \vspace{-10pt}
\end{figure}

To investigate the optimal choice of decoding length $ d $ in CFIC, we conducted experiments with various settings of this parameter. The results of these experiments are depicted in Figure \ref{fig:window}. Our findings substantiate the initial hypotheses: the performance across all tasks progressively improves and reaches its zenith at a $d$ value of 256. Beyond this point, performance begins to wane, suggesting that a setting of $d=256$ strikes an effective balance for these tasks. This observation aligns with the intuition that a span of 256 tokens typically suffices to encapsulate a semantically complete and coherent unit of information.

\paragraph{Case Study: CFIC v.s. GPTs}
 
OpenAI's model APIs, including GPT-3.5 and GPT-4, serve as robust baselines in the domain of LLM. However, they were excluded from the primary model comparisons in our experiments for two primary reasons:
(1). these APIs lack of control over decoding process resulting in the inability to manipulate their decoding mechanisms to align with our methodological requirements.
(2). The foundational models of GPT-3.5 and GPT-4 are characterized by their vast parameter sizes (\eg, 175 billion parameters), endowing them with exceptional language modeling capabilities, especially in handling extended contexts. However, our focus with CFIC is on applying LLMs with comparatively smaller parameter sizes. This approach ensures more manageable computational resource requirements and enhances model scalability.

Despite these exclusions, we conducted a comparative case study, the results of which are presented in Table \ref{tab:case}. This study reveals that our CFIC-7B model consistently provided complete and relevant grounding text evidence in response to queries. In contrast, the other models exhibited limitations:
(1). LongAlpaca-7B failed to accurately locate appropriate grounding text, resulting in the generation of information irrelevant to the downstream tasks.
(2). GPT-3.5 is able to directly respond to queries, it did not successfully identify precise grounding text from the original source material.
(3). Although GPT-4 managed to retrieve grounding text pertinent to the query, the information provided was incomplete, lacking the necessary comprehensiveness to fully support the response logically.

\section{Conclusion}
This study introduces a Chunking-Free In-Context (CFIC) retrieval method for the RAG system, addressing the challenges of processing lengthy documents and refining evidence retrieval. Unlike traditional chunking-based methods that either compromise textual coherence or struggle with noise and inaccuracies, CFIC leverages auto-aggressive decoding to pinpoint precise evidence directly, eliminating the reliance on chunking. CFIC incorporates Constrained Sentence Prefix Decoding and Skip Decoding strategies to further enhances retrieval efficiency and accuracy. Through comprehensive evaluations on various open QA datasets, CFIC has demonstrated remarkable improvements in sourcing relevant and precise evidence to ground language models.

\section*{Limitations}
This paper introduces a novel approach for Retrieval-Augmented Generation systems through the Chunking-Free In-Context (CFIC) retrieval method. Despite its advancements and effectiveness, there are certain limitations that warrant discussion.

One of the primary limitations stems from the training data used to develop our models. The dataset, self-constructed and annotated using ChatGPT, may harbor annotation biases. Such biases could affect the model's performance, particularly in its ability to generalize across different types of data or domains. While our approach excels in tasks requiring precise text evidence, it may offer limited assistance in scenarios demanding a high-level understanding of context, such as summarization tasks. This limitation is due to the model's focused capability on specific evidence retrieval rather than broader context comprehension.

Additionally, in this study, we have set the maximum length that CFIC can handle to 32k tokens. While this threshold accommodates a wide range of documents, it may not suffice for longer texts, such as novels, which exceed this limit. This constraint is primarily dictated by the available computational resources, highlighting a need for more efficient processing methods or greater computational power to extend CFIC's applicability to longer documents.
With the increase in computational resources and advancements in model acceleration algorithms, we envision the future possibility of enabling CFIC to handle even longer contexts. This could potentially extend to encoding the entire corpus, facilitating corpus-level in-context retrieval for each query.

\section*{Ethical Impact}
The development of CFIC builds upon existing Large Language Models (LLMs), which are trained on vast, diverse text corpora. This foundation introduces potential risks associated with biases inherent in the original training data. These biases can manifest in the model's outputs, influencing the quality and impartiality of the retrieved evidence.

Furthermore, the long documents processed by CFIC are sourced from the web, a domain rife with its biases. The web's text content reflects a wide array of perspectives, some of which may be skewed or unrepresentative of broader viewpoints. Given that CFIC is designed to process and retrieve information from these documents, there is a risk that the model might inadvertently perpetuate or amplify these biases without the capacity to discern or mitigate them.

Addressing these limitations and potential risks is crucial for the continued development and refinement of CFIC. Future work could explore more robust training datasets, advanced bias mitigation strategies, and enhanced processing capabilities to broaden the model's applicability and ensure the fairness and accuracy of its outputs.

\bibliography{anthology,custom}

\begin{thebibliography}{56}
\expandafter\ifx\csname natexlab\endcsname\relax\def\natexlab#1{#1}\fi

\bibitem[{Asai et~al.(2022)Asai, Gardner, and Hajishirzi}]{asai-etal-2022-evidentiality}
Akari Asai, Matt Gardner, and Hannaneh Hajishirzi. 2022.
\newblock \href {https://doi.org/10.18653/v1/2022.naacl-main.162} {Evidentiality-guided generation for knowledge-intensive {NLP} tasks}.
\newblock In \emph{Proceedings of the 2022 Conference of the North American Chapter of the Association for Computational Linguistics: Human Language Technologies}, pages 2226--2243, Seattle, United States. Association for Computational Linguistics.

\bibitem[{Bai et~al.(2023)Bai, Lv, Zhang, Lyu, Tang, Huang, Du, Liu, Zeng, Hou, Dong, Tang, and Li}]{bai2023longbench}
Yushi Bai, Xin Lv, Jiajie Zhang, Hongchang Lyu, Jiankai Tang, Zhidian Huang, Zhengxiao Du, Xiao Liu, Aohan Zeng, Lei Hou, Yuxiao Dong, Jie Tang, and Juanzi Li. 2023.
\newblock \href {http://arxiv.org/abs/2308.14508} {Longbench: A bilingual, multitask benchmark for long context understanding}.

\bibitem[{Bang et~al.(2023)Bang, Cahyawijaya, Lee, Dai, Su, Wilie, Lovenia, Ji, Yu, Chung et~al.}]{multitaskhallu}
Yejin Bang, Samuel Cahyawijaya, Nayeon Lee, Wenliang Dai, Dan Su, Bryan Wilie, Holy Lovenia, Ziwei Ji, Tiezheng Yu, Willy Chung, et~al. 2023.
\newblock A multitask, multilingual, multimodal evaluation of chatgpt on reasoning, hallucination, and interactivity.
\newblock \emph{arXiv preprint arXiv:2302.04023}.

\bibitem[{Brown et~al.(2020)Brown, Mann, Ryder, Subbiah, Kaplan, Dhariwal, Neelakantan, Shyam, Sastry, Askell et~al.}]{gpt-3}
Tom Brown, Benjamin Mann, Nick Ryder, Melanie Subbiah, Jared~D Kaplan, Prafulla Dhariwal, Arvind Neelakantan, Pranav Shyam, Girish Sastry, Amanda Askell, et~al. 2020.
\newblock Language models are few-shot learners.
\newblock \emph{Advances in neural information processing systems}, 33:1877--1901.

\bibitem[{Chen et~al.(2023)Chen, Qian, Tang, Lai, Liu, Han, and Jia}]{longlora}
Yukang Chen, Shengju Qian, Haotian Tang, Xin Lai, Zhijian Liu, Song Han, and Jiaya Jia. 2023.
\newblock Longlora: Efficient fine-tuning of long-context large language models.
\newblock \emph{arXiv:2309.12307}.

\bibitem[{Chung et~al.(2022)Chung, Hou, Longpre, Zoph, Tay, Fedus, Li, Wang, Dehghani, Brahma, Webson, Gu, Dai, Suzgun, Chen, Chowdhery, Castro-Ros, Pellat, Robinson, Valter, Narang, Mishra, Yu, Zhao, Huang, Dai, Yu, Petrov, Chi, Dean, Devlin, Roberts, Zhou, Le, and Wei}]{chung2022scaling}
Hyung~Won Chung, Le~Hou, Shayne Longpre, Barret Zoph, Yi~Tay, William Fedus, Yunxuan Li, Xuezhi Wang, Mostafa Dehghani, Siddhartha Brahma, Albert Webson, Shixiang~Shane Gu, Zhuyun Dai, Mirac Suzgun, Xinyun Chen, Aakanksha Chowdhery, Alex Castro-Ros, Marie Pellat, Kevin Robinson, Dasha Valter, Sharan Narang, Gaurav Mishra, Adams Yu, Vincent Zhao, Yanping Huang, Andrew Dai, Hongkun Yu, Slav Petrov, Ed~H. Chi, Jeff Dean, Jacob Devlin, Adam Roberts, Denny Zhou, Quoc~V. Le, and Jason Wei. 2022.
\newblock \href {https://arxiv.org/pdf/2210.11416} {Scaling instruction-finetuned language models}.
\newblock \emph{arXiv preprint arXiv:2210.11416}.

\bibitem[{Dasigi et~al.(2021)Dasigi, Lo, Beltagy, Cohan, Smith, and Gardner}]{dasigi2021dataset}
Pradeep Dasigi, Kyle Lo, Iz~Beltagy, Arman Cohan, Noah~A. Smith, and Matt Gardner. 2021.
\newblock \href {http://arxiv.org/abs/2105.03011} {A dataset of information-seeking questions and answers anchored in research papers}.

\bibitem[{Dong et~al.(2023)Dong, Tang, Li, and Zhao}]{dong2023survey}
Zican Dong, Tianyi Tang, Lunyi Li, and Wayne~Xin Zhao. 2023.
\newblock A survey on long text modeling with transformers.
\newblock \emph{arXiv preprint arXiv:2302.14502}.

\bibitem[{Gao et~al.(2024)Gao, Xiong, Gao, Jia, Pan, Bi, Dai, Sun, Guo, Wang, and Wang}]{gao2024retrievalaugmented}
Yunfan Gao, Yun Xiong, Xinyu Gao, Kangxiang Jia, Jinliu Pan, Yuxi Bi, Yi~Dai, Jiawei Sun, Qianyu Guo, Meng Wang, and Haofen Wang. 2024.
\newblock \href {http://arxiv.org/abs/2312.10997} {Retrieval-augmented generation for large language models: A survey}.

\bibitem[{Google(2023)}]{gemini}
Google. 2023.
\newblock Gemini: A family of highly capable multimodal models.
\newblock \url{https://goo.gle/GeminiPaper}.

\bibitem[{Guo et~al.(2022)Guo, Ma, Mao, Qian, Zhang, Jiang, Cao, and Dou}]{guo22webformer}
Yu~Guo, Zhengyi Ma, Jiaxin Mao, Hongjin Qian, Xinyu Zhang, Hao Jiang, Zhao Cao, and Zhicheng Dou. 2022.
\newblock \href {https://doi.org/10.1145/3477495.3532086} {Webformer: Pre-training with web pages for information retrieval}.
\newblock In \emph{Proceedings of the 45th International ACM SIGIR Conference on Research and Development in Information Retrieval}, SIGIR '22, page 1502–1512, New York, NY, USA. Association for Computing Machinery.

\bibitem[{Guu et~al.(2020)Guu, Lee, Tung, Pasupat, and Chang}]{guu2020realm}
Kelvin Guu, Kenton Lee, Zora Tung, Panupong Pasupat, and Ming-Wei Chang. 2020.
\newblock \href {https://dl.acm.org/doi/abs/10.5555/3524938.3525306} {{REALM}: Retrieval-augmented language model pre-training}.
\newblock In \emph{International Conference on Machine Learning}. JMLR.org.

\bibitem[{Izacard and Grave(2020)}]{FiD}
Gautier Izacard and Edouard Grave. 2020.
\newblock Distilling knowledge from reader to retriever for question answering.
\newblock \emph{arXiv preprint arXiv:2012.04584}.

\bibitem[{Izacard and Grave(2021{\natexlab{a}})}]{izacard2021distilling}
Gautier Izacard and Edouard Grave. 2021{\natexlab{a}}.
\newblock \href {https://openreview.net/forum?id=NTEz-6wysdb} {Distilling knowledge from reader to retriever for question answering}.
\newblock In \emph{International Conference on Learning Representations}.

\bibitem[{Izacard and Grave(2021{\natexlab{b}})}]{izacard2021leveraging}
Gautier Izacard and Edouard Grave. 2021{\natexlab{b}}.
\newblock \href {http://arxiv.org/abs/2007.01282} {Leveraging passage retrieval with generative models for open domain question answering}.

\bibitem[{Ji et~al.(2023)Ji, Lee, Frieske, Yu, Su, Xu, Ishii, Bang, Madotto, and Fung}]{ji2023survey}
Ziwei Ji, Nayeon Lee, Rita Frieske, Tiezheng Yu, Dan Su, Yan Xu, Etsuko Ishii, Ye~Jin Bang, Andrea Madotto, and Pascale Fung. 2023.
\newblock Survey of hallucination in natural language generation.
\newblock \emph{ACM Computing Surveys}, 55(12):1--38.

\bibitem[{Jiang et~al.(2023)Jiang, Xu, Gao, Sun, Liu, Dwivedi-Yu, Yang, Callan, and Neubig}]{jiang2023active}
Zhengbao Jiang, Frank~F. Xu, Luyu Gao, Zhiqing Sun, Qian Liu, Jane Dwivedi-Yu, Yiming Yang, Jamie Callan, and Graham Neubig. 2023.
\newblock \href {https://arxiv.org/pdf/2305.06983} {Active retrieval augmented generation}.
\newblock \emph{arXiv preprint arXiv:2305.06983}.

\bibitem[{Kaddour et~al.(2023)Kaddour, Harris, Mozes, Bradley, Raileanu, and McHardy}]{kaddour2023challenges}
Jean Kaddour, Joshua Harris, Maximilian Mozes, Herbie Bradley, Roberta Raileanu, and Robert McHardy. 2023.
\newblock Challenges and applications of large language models.
\newblock \emph{arXiv preprint arXiv:2307.10169}.

\bibitem[{Kamalloo et~al.(2023)Kamalloo, Dziri, Clarke, and Rafiei}]{kamalloo2023evaluating}
Ehsan Kamalloo, Nouha Dziri, Charles~LA Clarke, and Davood Rafiei. 2023.
\newblock \href {https://arxiv.org/pdf/2305.06984} {Evaluating open-domain question answering in the era of large language models}.
\newblock \emph{arXiv preprint arXiv:2305.06984}.

\bibitem[{Kang et~al.(2023)Kang, Kwak, Baek, and Hwang}]{sugre}
Minki Kang, Jin~Myung Kwak, Jinheon Baek, and Sung~Ju Hwang. 2023.
\newblock Knowledge graph-augmented language models for knowledge-grounded dialogue generation.
\newblock \emph{arXiv preprint arXiv:2305.18846}.

\bibitem[{Kaplan et~al.(2020)Kaplan, McCandlish, Henighan, Brown, Chess, Child, Gray, Radford, Wu, and Amodei}]{kaplan2020scaling}
Jared Kaplan, Sam McCandlish, Tom Henighan, Tom~B. Brown, Benjamin Chess, Rewon Child, Scott Gray, Alec Radford, Jeffrey Wu, and Dario Amodei. 2020.
\newblock \href {http://arxiv.org/abs/2001.08361} {Scaling laws for neural language models}.

\bibitem[{Karpukhin et~al.(2020)Karpukhin, Oguz, Min, Lewis, Wu, Edunov, Chen, and Yih}]{karpukhin-etal-2020-dense}
Vladimir Karpukhin, Barlas Oguz, Sewon Min, Patrick Lewis, Ledell Wu, Sergey Edunov, Danqi Chen, and Wen-tau Yih. 2020.
\newblock \href {https://doi.org/10.18653/v1/2020.emnlp-main.550} {Dense passage retrieval for open-domain question answering}.
\newblock In \emph{Proceedings of the 2020 Conference on Empirical Methods in Natural Language Processing (EMNLP)}, pages 6769--6781, Online. Association for Computational Linguistics.

\bibitem[{Khandelwal et~al.(2019)Khandelwal, Levy, Jurafsky, Zettlemoyer, and Lewis}]{kNN-LM}
Urvashi Khandelwal, Omer Levy, Dan Jurafsky, Luke Zettlemoyer, and Mike Lewis. 2019.
\newblock Generalization through memorization: Nearest neighbor language models.
\newblock \emph{arXiv preprint arXiv:1911.00172}.

\bibitem[{Komeili et~al.(2022)Komeili, Shuster, and Weston}]{komeili-etal-2022-internet}
Mojtaba Komeili, Kurt Shuster, and Jason Weston. 2022.
\newblock \href {https://doi.org/10.18653/v1/2022.acl-long.579} {{I}nternet-augmented dialogue generation}.
\newblock In \emph{Proceedings of the 60th Annual Meeting of the Association for Computational Linguistics (Volume 1: Long Papers)}, pages 8460--8478, Dublin, Ireland. Association for Computational Linguistics.

\bibitem[{Kočiský et~al.(2017)Kočiský, Schwarz, Blunsom, Dyer, Hermann, Melis, and Grefenstette}]{kočiský2017narrativeqa}
Tomáš Kočiský, Jonathan Schwarz, Phil Blunsom, Chris Dyer, Karl~Moritz Hermann, Gábor Melis, and Edward Grefenstette. 2017.
\newblock \href {http://arxiv.org/abs/1712.07040} {The narrativeqa reading comprehension challenge}.

\bibitem[{Lewis et~al.(2020)Lewis, Perez, Piktus, Petroni, Karpukhin, Goyal, K\"{u}ttler, Lewis, Yih, Rockt\"{a}schel, Riedel, and Kiela}]{lewis2020retrieval}
Patrick Lewis, Ethan Perez, Aleksandra Piktus, Fabio Petroni, Vladimir Karpukhin, Naman Goyal, Heinrich K\"{u}ttler, Mike Lewis, Wen-tau Yih, Tim Rockt\"{a}schel, Sebastian Riedel, and Douwe Kiela. 2020.
\newblock \href {https://proceedings.neurips.cc/paper_files/paper/2020/file/6b493230205f780e1bc26945df7481e5-Paper.pdf} {{R}etrieval-{A}ugmented {G}eneration for knowledge-intensive {NLP} tasks}.
\newblock In \emph{Advances in Neural Information Processing Systems}, volume~33, pages 9459--9474.

\bibitem[{Li et~al.(2023)Li, Shao, Xie, Sheng, Zheng, Gonzalez, Stoica, Ma, and Zhang}]{longchat2023}
Dacheng Li, Rulin Shao, Anze Xie, Ying Sheng, Lianmin Zheng, Joseph~E. Gonzalez, Ion Stoica, Xuezhe Ma, and Hao Zhang. 2023.
\newblock \href {https://lmsys.org/blog/2023-06-29-longchat} {How long can open-source llms truly promise on context length?}

\bibitem[{Li et~al.(2022{\natexlab{a}})Li, Rawat, Zaheer, Wang, Lukasik, Veit, Yu, and Kumar}]{li2022large}
Daliang Li, Ankit~Singh Rawat, Manzil Zaheer, Xin Wang, Michal Lukasik, Andreas Veit, Felix Yu, and Sanjiv Kumar. 2022{\natexlab{a}}.
\newblock \href {http://arxiv.org/abs/2211.05110} {Large language models with controllable working memory}.

\bibitem[{Li et~al.(2022{\natexlab{b}})Li, Wu, Chen, Liu, Xiao, and Wu}]{li2022faithfulness}
Wei Li, Wenhao Wu, Moye Chen, Jiachen Liu, Xinyan Xiao, and Hua Wu. 2022{\natexlab{b}}.
\newblock Faithfulness in natural language generation: A systematic survey of analysis, evaluation and optimization methods.
\newblock \emph{arXiv preprint arXiv:2203.05227}.

\bibitem[{Liu et~al.(2023{\natexlab{a}})Liu, Lin, Hewitt, Paranjape, Bevilacqua, Petroni, and Liang}]{liu2023lost}
Nelson~F. Liu, Kevin Lin, John Hewitt, Ashwin Paranjape, Michele Bevilacqua, Fabio Petroni, and Percy Liang. 2023{\natexlab{a}}.
\newblock \href {http://arxiv.org/abs/2307.03172} {Lost in the middle: How language models use long contexts}.

\bibitem[{Liu et~al.(2023{\natexlab{b}})Liu, Lai, Yu, Xu, Zeng, Du, Zhang, Dong, and Tang}]{WebGLM}
Xiao Liu, Hanyu Lai, Hao Yu, Yifan Xu, Aohan Zeng, Zhengxiao Du, Peng Zhang, Yuxiao Dong, and Jie Tang. 2023{\natexlab{b}}.
\newblock Webglm: Towards an efficient web-enhanced question answering system with human preferences.
\newblock \emph{arXiv preprint arXiv:2306.07906}.

\bibitem[{Mao et~al.(2023)Mao, Dou, Mo, Hou, Chen, and Qian}]{mao2023large}
Kelong Mao, Zhicheng Dou, Fengran Mo, Jiewen Hou, Haonan Chen, and Hongjin Qian. 2023.
\newblock \href {http://arxiv.org/abs/2303.06573} {Large language models know your contextual search intent: A prompting framework for conversational search}.

\bibitem[{Mao et~al.(2021)Mao, He, Liu, Shen, Gao, Han, and Chen}]{mao-etal-2021-reader}
Yuning Mao, Pengcheng He, Xiaodong Liu, Yelong Shen, Jianfeng Gao, Jiawei Han, and Weizhu Chen. 2021.
\newblock \href {https://doi.org/10.18653/v1/2021.findings-acl.29} {Reader-guided passage reranking for open-domain question answering}.
\newblock In \emph{Findings of the Association for Computational Linguistics: ACL-IJCNLP 2021}, pages 344--350, Online. Association for Computational Linguistics.

\bibitem[{Nishikawa et~al.(2022)Nishikawa, Ri, Yamada, Tsuruoka, and Echizen}]{ease}
Sosuke Nishikawa, Ryokan Ri, Ikuya Yamada, Yoshimasa Tsuruoka, and Isao Echizen. 2022.
\newblock Ease: Entity-aware contrastive learning of sentence embedding.
\newblock \emph{arXiv preprint arXiv:2205.04260}.

\bibitem[{Nogueira and Cho(2020)}]{nogueira2020passage}
Rodrigo Nogueira and Kyunghyun Cho. 2020.
\newblock \href {https://arxiv.org/pdf/1901.04085} {Passage re-ranking with bert}.
\newblock \emph{arXiv preprint arXiv:1901.04085}.

\bibitem[{OpenAI(2023)}]{gpt-4}
OpenAI. 2023.
\newblock Gpt-4 technical report.
\newblock \url{https://cdn.openai.com/papers/gpt-4.pdf}.

\bibitem[{Qian et~al.(2023{\natexlab{a}})Qian, Dou, Tan, Chen, Gu, Lai, Zhang, Cao, and Wen}]{qian2023optimizing}
Hongjin Qian, Zhicheng Dou, Jiejun Tan, Haonan Chen, Haoqi Gu, Ruofei Lai, Xinyu Zhang, Zhao Cao, and Ji-Rong Wen. 2023{\natexlab{a}}.
\newblock \href {http://arxiv.org/abs/2308.15711} {Optimizing factual accuracy in text generation through dynamic knowledge selection}.

\bibitem[{Qian et~al.(2023{\natexlab{b}})Qian, Zhu, Dou, Gu, Zhang, Liu, Lai, Cao, Nie, and Wen}]{qian2023webbrain}
Hongjing Qian, Yutao Zhu, Zhicheng Dou, Haoqi Gu, Xinyu Zhang, Zheng Liu, Ruofei Lai, Zhao Cao, Jian-Yun Nie, and Ji-Rong Wen. 2023{\natexlab{b}}.
\newblock \href {http://arxiv.org/abs/2304.04358} {Webbrain: Learning to generate factually correct articles for queries by grounding on large web corpus}.

\bibitem[{Ram et~al.(2023)Ram, Levine, Dalmedigos, Muhlgay, Shashua, Leyton-Brown, and Shoham}]{ICRALM}
Ori Ram, Yoav Levine, Itay Dalmedigos, Dor Muhlgay, Amnon Shashua, Kevin Leyton-Brown, and Yoav Shoham. 2023.
\newblock In-context retrieval-augmented language models.
\newblock \emph{arXiv preprint arXiv:2302.00083}.

\bibitem[{Ratner et~al.(2022)Ratner, Levine, Belinkov, Ram, Magar, Abend, Karpas, Shashua, Leyton-Brown, and Shoham}]{pcw}
Nir Ratner, Yoav Levine, Yonatan Belinkov, Ori Ram, Inbal Magar, Omri Abend, Ehud Karpas, Amnon Shashua, Kevin Leyton-Brown, and Yoav Shoham. 2022.
\newblock \href {https://doi.org/10.48550/arxiv.2212.10947} {{Parallel Context Windows Improve In-Context Learning of Large Language Models}}.
\newblock \emph{arXiv}.
\newblock Window.

\bibitem[{Shi et~al.(2023)Shi, Chen, Misra, Scales, Dohan, Chi, Schärli, and Zhou}]{shi2023large}
Freda Shi, Xinyun Chen, Kanishka Misra, Nathan Scales, David Dohan, Ed~Chi, Nathanael Schärli, and Denny Zhou. 2023.
\newblock \href {https://arxiv.org/pdf/2302.00093} {Large language models can be easily distracted by irrelevant context}.
\newblock \emph{arXiv preprint arXiv:2302.00093}.

\bibitem[{Shuster et~al.(2021{\natexlab{a}})Shuster, Poff, Chen, Kiela, and Weston}]{shuster-etal-2021-retrieval-augmentation}
Kurt Shuster, Spencer Poff, Moya Chen, Douwe Kiela, and Jason Weston. 2021{\natexlab{a}}.
\newblock \href {https://doi.org/10.18653/v1/2021.findings-emnlp.320} {Retrieval augmentation reduces hallucination in conversation}.
\newblock In \emph{Findings of the Association for Computational Linguistics: EMNLP 2021}, pages 3784--3803, Punta Cana, Dominican Republic. Association for Computational Linguistics.

\bibitem[{Shuster et~al.(2021{\natexlab{b}})Shuster, Poff, Chen, Kiela, and Weston}]{raghallucination}
Kurt Shuster, Spencer Poff, Moya Chen, Douwe Kiela, and Jason Weston. 2021{\natexlab{b}}.
\newblock Retrieval augmentation reduces hallucination in conversation.
\newblock \emph{arXiv preprint arXiv:2104.07567}.

\bibitem[{Touvron et~al.(2023{\natexlab{a}})Touvron, Martin, Stone, Albert, Almahairi, Babaei, Bashlykov, Batra, Bhargava, Bhosale et~al.}]{llama}
Hugo Touvron, Louis Martin, Kevin Stone, Peter Albert, Amjad Almahairi, Yasmine Babaei, Nikolay Bashlykov, Soumya Batra, Prajjwal Bhargava, Shruti Bhosale, et~al. 2023{\natexlab{a}}.
\newblock Llama 2: Open foundation and fine-tuned chat models.
\newblock \emph{arXiv preprint arXiv:2307.09288}.

\bibitem[{Touvron et~al.(2023{\natexlab{b}})Touvron, Martin, Stone, Albert, Almahairi, Babaei, Bashlykov, Batra, Bhargava, Bhosale et~al.}]{touvron2023llama}
Hugo Touvron, Louis Martin, Kevin Stone, Peter Albert, Amjad Almahairi, Yasmine Babaei, Nikolay Bashlykov, Soumya Batra, Prajjwal Bhargava, Shruti Bhosale, et~al. 2023{\natexlab{b}}.
\newblock \href {https://arxiv.org/pdf/2307.09288} {Llama 2: Open foundation and fine-tuned chat models}.
\newblock \emph{arXiv preprint arXiv:2307.09288}.

\bibitem[{Trivedi et~al.(2022)Trivedi, Balasubramanian, Khot, and Sabharwal}]{trivedi2022musique}
Harsh Trivedi, Niranjan Balasubramanian, Tushar Khot, and Ashish Sabharwal. 2022.
\newblock \href {http://arxiv.org/abs/2108.00573} {Musique: Multihop questions via single-hop question composition}.

\bibitem[{Wang et~al.(2019)Wang, Ng, Ma, Nallapati, and Xiang}]{wang-etal-2019-multi}
Zhiguo Wang, Patrick Ng, Xiaofei Ma, Ramesh Nallapati, and Bing Xiang. 2019.
\newblock \href {https://doi.org/10.18653/v1/D19-1599} {Multi-passage {BERT}: A globally normalized {BERT} model for open-domain question answering}.
\newblock In \emph{Proceedings of the 2019 Conference on Empirical Methods in Natural Language Processing and the 9th International Joint Conference on Natural Language Processing (EMNLP-IJCNLP)}, pages 5878--5882, Hong Kong, China. Association for Computational Linguistics.

\bibitem[{Xiao et~al.(2023)Xiao, Liu, Zhang, and Muennighoff}]{bge_embedding}
Shitao Xiao, Zheng Liu, Peitian Zhang, and Niklas Muennighoff. 2023.
\newblock \href {http://arxiv.org/abs/2309.07597} {C-pack: Packaged resources to advance general chinese embedding}.

\bibitem[{Xu et~al.(2023{\natexlab{a}})Xu, Ping, Wu, McAfee, Zhu, Liu, Subramanian, Bakhturina, Shoeybi, and Catanzaro}]{longctx}
Peng Xu, Wei Ping, Xianchao Wu, Lawrence McAfee, Chen Zhu, Zihan Liu, Sandeep Subramanian, Evelina Bakhturina, Mohammad Shoeybi, and Bryan Catanzaro. 2023{\natexlab{a}}.
\newblock \href {https://doi.org/10.48550/arxiv.2310.03025} {{Retrieval meets Long Context Large Language Models}}.
\newblock \emph{arXiv}.
\newblock Experimental.

\bibitem[{Xu et~al.(2023{\natexlab{b}})Xu, Ping, Wu, McAfee, Zhu, Liu, Subramanian, Bakhturina, Shoeybi, and Catanzaro}]{xu2023retrieval}
Peng Xu, Wei Ping, Xianchao Wu, Lawrence McAfee, Chen Zhu, Zihan Liu, Sandeep Subramanian, Evelina Bakhturina, Mohammad Shoeybi, and Bryan Catanzaro. 2023{\natexlab{b}}.
\newblock Retrieval meets long context large language models.
\newblock \emph{arXiv preprint arXiv:2310.03025}.

\bibitem[{Yang et~al.(2023)Yang, Yue, and He}]{AutoGPT}
Hui Yang, Sifu Yue, and Yunzhong He. 2023.
\newblock Auto-gpt for online decision making: Benchmarks and additional opinions.
\newblock \emph{arXiv preprint arXiv:2306.02224}.

\bibitem[{Yang et~al.(2018)Yang, Qi, Zhang, Bengio, Cohen, Salakhutdinov, and Manning}]{yang2018hotpotqa}
Zhilin Yang, Peng Qi, Saizheng Zhang, Yoshua Bengio, William~W. Cohen, Ruslan Salakhutdinov, and Christopher~D. Manning. 2018.
\newblock \href {http://arxiv.org/abs/1809.09600} {Hotpotqa: A dataset for diverse, explainable multi-hop question answering}.

\bibitem[{Yao et~al.(2023)Yao, Ning, Liu, Ning, and Yuan}]{llmlie}
Jia-Yu Yao, Kun-Peng Ning, Zhen-Hui Liu, Mu-Nan Ning, and Li~Yuan. 2023.
\newblock Llm lies: Hallucinations are not bugs, but features as adversarial examples.
\newblock \emph{arXiv preprint arXiv:2310.01469}.

\bibitem[{Zhang et~al.(2023{\natexlab{a}})Zhang, Xiao, Liu, Dou, and Nie}]{llm_embedder}
Peitian Zhang, Shitao Xiao, Zheng Liu, Zhicheng Dou, and Jian-Yun Nie. 2023{\natexlab{a}}.
\newblock \href {http://arxiv.org/abs/2310.07554} {Retrieve anything to augment large language models}.

\bibitem[{Zhang et~al.(2023{\natexlab{b}})Zhang, Li, Cui, Cai, Liu, Fu, Huang, Zhao, Zhang, Chen et~al.}]{hallucination}
Yue Zhang, Yafu Li, Leyang Cui, Deng Cai, Lemao Liu, Tingchen Fu, Xinting Huang, Enbo Zhao, Yu~Zhang, Yulong Chen, et~al. 2023{\natexlab{b}}.
\newblock Siren's song in the ai ocean: A survey on hallucination in large language models.
\newblock \emph{arXiv preprint arXiv:2309.01219}.

\bibitem[{Zheng et~al.(2023)Zheng, Chiang, Sheng, Zhuang, Wu, Zhuang, Lin, Li, Li, Xing, Zhang, Gonzalez, and Stoica}]{zheng2023judging}
Lianmin Zheng, Wei-Lin Chiang, Ying Sheng, Siyuan Zhuang, Zhanghao Wu, Yonghao Zhuang, Zi~Lin, Zhuohan Li, Dacheng Li, Eric.~P Xing, Hao Zhang, Joseph~E. Gonzalez, and Ion Stoica. 2023.
\newblock \href {http://arxiv.org/abs/2306.05685} {Judging llm-as-a-judge with mt-bench and chatbot arena}.

\end{thebibliography}
\bibliographystyle{acl_natbib}

\appendix

\section{Implementation Detail}
\label{sec:imp}
To train CFIC, we employed the ``LLAMA2-7B-chat'' as the foundation model for our CFIC. During the training, we set the batch size to 1 per GPU and the learning rate to 1e-5. We set the gradient accumulation step as 8 and utilized the AdamW optimizer with an epsilon value of 1e-8. The model's maximum length parameter was set to 32768. We train the model for 600 steps on 8 * Nvidia A800 80GB GPUs. For CFIC, We set the number of sampled sentence prefixes as $k=3$ and the maximum decoding length as $d=256$~(refers to Eq.~(\ref{eq:skip})). Besides, we use a warm-up strategy to adjust the learning rate. To save GPU memory, we employed DeepSpeed's Stage 2 zero optimization to save GPU memory.


\end{document}